\pdfoutput=1

\documentclass[11pt]{article}

\usepackage{acl}
\usepackage{times}
\usepackage{latexsym}
\usepackage{rotating}
\usepackage{mathtools}
\usepackage{xcolor}
\usepackage{soul}
\usepackage{booktabs}
\usepackage{multirow}
\usepackage{hyperref}

\usepackage[T1]{fontenc}

\usepackage[utf8]{inputenc}

\usepackage{microtype}

\title{Multi-Task Training with In-Domain Language Models for Diagnostic Reasoning}

\author{Brihat Sharma$^{1}$, Yanjun Gao$^{1}$, Timothy Miller$^{2}$, Matthew M. Churpek$^{1}$,\\
\textbf{Majid Afshar}$^{1}$ and \textbf{Dmitriy Dligach}$^{3}$ \\
$^{1}$University of Wisconsin-Madison, \\
$^{2}$Boston Children's Hospital and Harvard Medical School,
$^{3}$Loyola University Chicago \\ 
\texttt{$^{1}$bsharma25@wisc.edu,} \\
\texttt{$^{1}$\{ygao, mchurpek, mafshar\} @medicine.wisc.edu, }  \\ \texttt{$^{2}$Timothy.Miller@childrens.harvard.edu, $^{3}$ddligach@luc.edu}\\}

\newcommand{\drbench}{DR.BENCH~}
\newcommand{\drbenchsEND}{DR.BENCH}

\begin{document}
\maketitle
\begin{abstract}
\textit{Generative artificial intelligence (AI) is a promising direction for augmenting clinical diagnostic decision support and reducing diagnostic errors, a leading contributor to medical errors. 
To further the development of clinical AI systems, the Diagnostic Reasoning Benchmark (\drbenchsEND) was introduced as a comprehensive generative AI framework, comprised of six tasks representing key components in clinical reasoning. 
We present a comparative analysis of in-domain versus out-of-domain language models as well as multi-task versus single task training with a focus on the problem summarization task in \drbench \cite{GAO2023104286}. 
We demonstrate that a multi-task, clinically-trained language model outperforms its general domain counterpart by a large margin, establishing a new state-of-the-art performance, with a ROUGE-L score of 28.55.
This research underscores the value of domain-specific training for optimizing clinical diagnostic reasoning tasks.}

\end{abstract}

\section{Introduction}

The electronic health record (EHR) contains daily progress notes authored by healthcare providers to represent the daily changes in care plans for their patients, including an updated list of active diagnoses. The daily progress note is one of the most important note types in the EHR and contains the daily subjective and objective details in the patient's care, which is summarized into an assessment of the overall leading diagnoses with a treatment plan section \cite{gao-etal-2022-hierarchical}. However, note bloat is a common phenomenon in medical documentation intermixed with billing requirements, non-diagnostic information, and copy and paste from prior notes \cite{redundancy-notes}. These additional documentation practices contribute to provider burnout and cognitive overload \cite{burnout}. Problem-based charting is important to improve care throughput and help reduce diagnostic errors \cite{electronic-problem}. 

The medical reasoning process is complex and incorporates medical knowledge representation with analytical and experiential knowledge \cite{doi:10.1056/NEJMra054782}. \citeauthor{ForwardReasoning} developed a theory from the AI literature that experts use "forward-reasoning" from data to diagnosis \citeyear{ForwardReasoning}.
The recently released benchmark \drbench (Diagnostic Reasoning Benchmark) is intended to assess the ability of AI models to perform such reasoning, with multiple component tasks including diagnostic reasoning with EHR data for experiential knowledge, medical exams for knowledge representation, progress note structure prediction, and problem summarization tasks that included both extractive and abstractive medical diagnoses \cite{GAO2023104286}.

\begin{figure}
    \centering
    \small
    \includegraphics[width=\columnwidth]{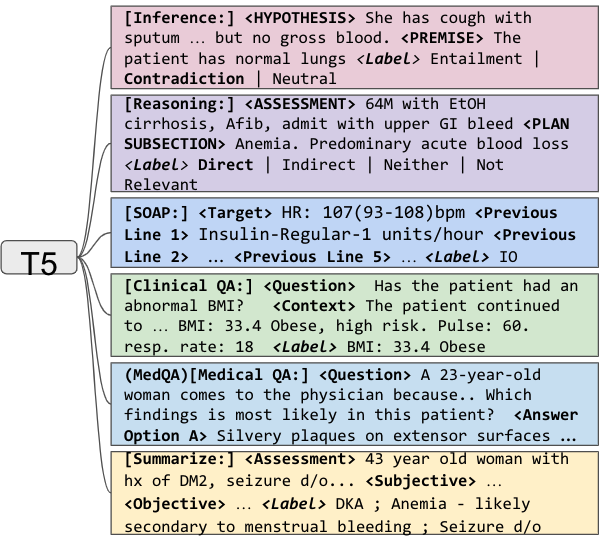}
    \vspace{-.25in}
    \caption{Training T5 with multi-task setup with six tasks from DR.BENCH~\cite{GAO2023104286}}
    \label{fig:multi_drbench}
\end{figure}
In this work, we focus primarily on the problem summarization task from the \drbench suite, but with the hypothesis that using all tasks in \drbench would improve the problem summarization task over the problem summarization task being fine-tuned alone. 
%
We make use of the T5 family of sequence-to-sequence language models,~\cite{limitsoftrans}, which are first pretrained on a large unlabeled dataset and then finetuned on specific multiple downstream tasks.    
%
%
The text-to-text approach in our experiment makes it possible to perform multi-task training. Hence, the T5 models were ideal for experimenting with single and multi-task techniques. 

Further, we experimented with a recently developed clinically-trained T5 model to quantify the value of in-domain pretraining data \cite{ClinicalT5}. We make our software publically available at \href{https://git.doit.wisc.edu/smph-public/dom/uw-icu-data-science-lab-public/drbench}{https://git.doit.wisc.edu/smph-public/dom/uw-icu-data-science-lab-public/drbench}.

\begin{table*}
\centering
\captionsetup{justification=centering}
\begin{tabular}{p{3cm} p{5cm} p{2cm} p{4cm}}

\toprule
Model & Training Corpus  & Initialization & Citation  \\

\midrule
\small \textsc{T5 220M} & 
\multirow{2}{*}{\small Colossal Clean Crawled Corpus (C4)} & \small Random & \multirow{2}{*}{\small \cite{limitsoftrans}}\\

\small \textsc{T5 770M}  &  & \small Random \\
\midrule

\small \textsc{SciFive 220M} & \multirow{2}{*}{\small C4 + PubMed (abstracts) + PMC} & \small T5 220M & \multirow{2}{*}{\small \cite{phan2021scifive}}\\

\small \textsc{SciFive 770M} &  & \small T5 770M\\

\midrule

\small \textsc{Clinical-T5 220M}  & \multirow{2}{*}{\small MIMIC-III + MIMIC-IV} & \small T5 220M & \multirow{2}{*}{\small \cite{ClinicalT5}}\\

\small \textsc{Clinical-T5 770M} &  & \small Random\\

\bottomrule
\end{tabular}
\vspace{-.1in}
\caption{ T5 pretrained models used in the experiments. \footnotemark}
\label{tab:ModelOverview}
\end{table*}

\footnotetext{PubMed is a large open source biomedical and lifescience database consists of 35 million citation and abstract, and PMC (PubMed Central) consists of full articles. MIMIC-III and MIMIC-IV (Medical Information Mart for Intensive Care) are databases consisting of de-identified datasets from Beth Israel Deaconess Medical Center}

\section{Related Work}

In the clinical domain, biomedical text summarization is a growing field. 
Common approaches to text summarization include feature-based methods ~\cite{PATEL2019167}, fine-tuning large language models ~\cite{lewis-etal-2020-bart}, and domain adaptation with fine-tuning methods~\cite{xie2023survey}. Researchers developed clinical methods for summarization from progress notes but these methods were restricted to specific diseases such as diabetes and hypertension~\cite{liang-etal-2019-novel-system}. Moreover, these methods for summarization were more extractive than abstractive, using a combination of heuristics rules and deep learning techniques, and did not use large language models~\cite{liang-etal-2019-novel-system}. In another work, an extractive-abstractive approach was used where meaningful sentences were extracted from the clinical notes first; these sentences were then fed into the transformer model for abstractive summarization ~\cite{pilault-etal-2020-extractive}. Unfortunately, the transformer model frequently produced hallucinated outputs, and was not coherent when compared to the ground truth~\cite{pilault-etal-2020-extractive}. In a similar extractive-abstractive approach, researchers used a pointer generator network to generate a note summary cluster and a language model such as T5 to generate an abstractive summary~\cite{krishna-etal-2021-generating}. None of these approaches used multi-task training or focused on clinically trained encoder-decoder since clinical T5 was only recently introduced. Prior work has not addressed the challenge of abstractive reasoning, or they used a two-step process to create abstractions. Recently, researchers used domain adaptive T5 model trained on the biomedical dataset but did not experiment with multi-task settings~\cite{GAO2023104286}.

\section{Methods}

\subsection{Dataset}

In our experiments, we used \drbench \cite{GAO2023104286}, a recently introduced benchmark designed to evaluate diagnostic reasoning capabilities of generative language models. \drbench consists of three categories of tasks (two tasks per category), as shown in Figure~\ref{fig:multi_drbench}. From top to the bottom, the categories and six tasks are: \textbf{Medical Knowledge Representation:} (1) Medical Natural Language Inference (MedNLI) task that considered sentence pairs with the objective to determine whether the hypothesis sentence could be inferred from the premise sentence \cite{MedNLI} (14,049 sentence pairs total); (2) Assessment and Plan Reasoning (A\slash P) task whose objective was to label relations between the assessment and treatment plan sections (5,897 samples). \textbf{Clinical Evidence Understanding and Integration:} (1) Electronic Medical Records Question Answering (emrQA) whose objective was to answer questions based on discharge summaries (53,199 questions total) \cite{pampari-etal-2018-emrqa}; (2) Progress Note Section Labeling task whose objective was to labels SOAP sections in progress notes (134,089 samples) \cite{Task13}. \textbf{Diagnosis Generation and Summarization:} (1) Medical Board Exam Question Answering (MedQA) task that consisted of medical board exam question-answer pairs (12,725 pairs) \cite{app11146421-medqa}; (2) Problem Summarization (ProbSumm) task whose goal was to produce the list of relevant problems and diagnoses based on the input that consisted of the SOAP sections of progress notes (2,783 samples).

In this work, we focused primarily on the problem summarization task, which was the most difficult but also believed to be the most impactful of the six \drbench tasks for downstream clinical application.

\subsection{Experimental Setup}

In our experiments, we used six generative language models, all based on the Text-To-Text Transfer Transformer (T5) model \cite{limitsoftrans}. The text-to-text paradigm utilized by T5 was a natural choice for our stated goal of exploring multi-task learning: transforming T5 into a multi-task learner simply involved prefixing individual task instances with a task-specific prompt after which the model could be trained using the standard cross-entropy loss. 

Table \ref{tab:ModelOverview} provides details about the models. We compared a multi-task scenario in which T5 variants were fine-tuned on all \drbench tasks and a single-task scenario in which T5 was fine-tuned on the problem summarization task only. We trained T5 models as follows:

\textbf{Single-task training:} In single-task training for problem summarization, we used the text of the assessment, subjective and objective sections of the progress notes as input and trained T5 to generate the list of problems and diagnoses.

\textbf{Multi-task training:} In multi-task training, we combined all \drbench tasks into a single dataset and trained T5 to generate task-specific output given task-specific input. Training examples of each task were prefixed with a task-specific prompt. The open-book setting only was used for MedQA. The rest of preprocessing follows \cite{GAO2023104286}. 

To enable comparison with existing work \cite{GAO2023104286} we used ROUGE-L score \cite{lin-2004-rouge} as our evaluation metric. ROUGE-L uses the longest common subsequence statistics to compare model outputs. A resampling technique with 1000 bootstrap samples was used to estimate the 95\% confidence intervals (CI) \cite{10.2307/2246110-bootstrap}.

Note that the Clinical-T5 model used in our experiment was pretrained on the same data (MIMIC-III) that was annotated by some \drbench tasks (e.g. problem summarization and EmrQA). This setting is known as transductive learning. Trunsductive learning is a very realistic scenario for the clinical domain where due to privacy issues, language models are likely be pretrained on the data from the same institution as the data to which they would be applied. Obviously, it would also be interesting to investigate the performance of a T5 variant that was trained on a clinical corpus that was different from which the evaluation data were sourced. Unfortunately, this was not possible due to the fact that MIMIC was the only publicly available corpus of clinical notes and it was used for training clinical language models. 

The training data consisted of one progress note per unique patient. A separate cohort of unique patients was selected for the test set, ensuring no overlap between the train and test splits. All experiments used Adam optimizer with a learning rate of 1e-5, batch size of 8, beam size of 5, and 100 epochs with early stopping. The learning rate and batch size were picked based on the best hyperparameters found from the prior work \cite{GAO2023104286}. All experiments were completed on a single A100 GPU with 40 GB memory. The models were reviewed for error analysis by a critical care physician on the full test set of 86 progress notes and common observations were highlighted with examples in the error analysis. 

\section{Results and Discussion}

The results of our experiments are summarized in Table \ref{tab:submission6}. The full set of results including the confidence intervals is available in the Appendix (Table \ref{tab:submission4}).

Clinical-T5 770M trained in the multi-task setting demonstrated the best performance (28.55) for the Summarization task, establishing a new state-of-the art for this task. The single-task setting for the same T5 variant was a close second (28.28). 

T5 variants trained on in-domain data (SciFive and Clinical-T5) performed better than their general domain counterpart T5 models of the same size. All models, except Clinical-T5 experienced a drop in performance when trained in a multi-task approach. We hypothesize that the models pretrained on non-clinical data were overwhelmed with out-of-domain (i.e. clinical) data when trained in a multi-task way and failed to generalize as a result. Predictably, larger models performed at least as well as the smaller models and outperformed the smaller models in most scenarios.

Admittedly, our work leaves open the question of whether the state-of-the-art performance obtained by Clinical-T5 770M has to do with the fact that it was pretrained on MIMIC notes, which were also annotated in the problem summarization task. At the same time, the performance of other T5 variants, such as SciFive 770M, was close, without it pretraining on MIMIC. This suggests that another T5 variant trained on a corpus of clinical notes that was different from MIMIC would perform at least as well or better depending on the size of the pretraining corpus. It should be noted that the model of this size, 770M parameters, can very likely absorb significantly larger amounts of clinical notes than what was available in MIMIC~\cite{hoffmann2022training}. We leave verifying this hypothesis for future work. 

\begin{table}[ht]
\begin{tabular}{l l l}
\toprule
Model & Training & Summarization \\
\midrule
\small \citealp{GAO2023104286} & \small Single task & \small 7.60 (5.31 - 9.89)  \\
\midrule
\small \textsc{T5 220M} & \small Single task & \small 26.35 (22.18 - 30.52) \\
 & \small Multi-task & \small 24.84 (20.28 - 29.40) \\
\midrule
\small \textsc{T5 770M} & \small Single task & \small 26.90 (22.58 - 31.23) \\
 & \small Multi-task & \small 23.99 (19.86 - 28.13) \\
\midrule
\small \textsc{SciFive 220M} &\small Single task & \small 25.31 (21.45 - 29.17) \\
 & \small Multi-task & \small 24.38 (19.99 - 28.78) \\
\midrule
\small \textsc{SciFive 770M} & \small Single task & \small 27.31 (23.09 - 31.53) \\
 & \small Multi-task & \small 25.31 (21.45 - 29.17) \\
\midrule
\small \textsc{Clinical-T5} & \small Single task & \small 25.35 (21.19 - 29.51) \\
 \small \textsc{220M} & \small Multi-task & \small 26.21 (21.92 - 30.49) \\
\midrule
\small \textsc{Clinical-T5} & \small Single task & \small 28.28 (24.17 - 32.38) \\
 \small \textsc{770M}& \small Multi-task & \small \textbf{28.55} (24.29 - 32.80) \\
\hline

\end{tabular}
\vspace{-.12in}
\caption{Performance of fine-tuned T5 models on the summarization task. 95\% confidence intervals are included. The first row is a baseline representing the best performance on this task to date. Please see the Appendix for the full set of results.}
\label{tab:submission6}
\end{table}




\textbf{Error Analysis:} Although both clinical models produced similar ROUGE-L scores, the model trained in a single-task setting appeared to achieve better abstraction during error analysis. For the example in Table \ref{tab:erroranalysissnippit}, the assessment described sepsis but does not mention the source of the sepsis infection in multi-task Clinical-T5 770M. The data from the subjective and objective sections of the progress note described an abdominal source and lab results were consistent with a clostridium difficile infection. The multi-task prediction was able to generate sepsis but further generated text that the source was unclear. The single task performed better abstraction and generated clostridium difficile as the source for the infection, which was more accurate during expert review. In another diagnosis, the ground truth label was ``EtOH Withdrawal" (alcohol withdrawal). The multitask extracted ``altered mental status, hypertensive, tachycardia," (symptoms of withdrawal) whereas the single task was able to abstract ``DTs EtOH w d," (delirium tremens alcohol withdrawal - a type of severe alcohol withdrawal in critically ill patients). Again, the single task achieved greater accuracy with abstraction from symptoms of alcohol withdrawal presented in the earlier sections of the note.

\textbf{Resource Utilization:} The experiments were conducted on the Google Cloud Platform using one A100 40 GB NVIDIA GPU on a Linux base system. For all experiments, the total training time was approximately 250 hours for both single-task and multi-task approaches. The carbon emission footprint was 35.5 kilograms (kg) of CO2. However, the total carbon emission was only 4.5 kg of CO2 for the single-task experiments. \cite{lacoste2019quantifying}

\section{Conclusion}

In this work we experiment with the \drbench suite of tasks and established a new state-of-the-art result on the problem list generation task, a task critical for AI-assisted diagnostic reasoning. Our other contribution indicates that multi-task learning does not work well, unless in-domain data was used for pretraining and that included (unlabeled) task data during pretraining (a scenario known as transductive learning) leads to the best performance. Finally, our work provides evidence that generative models benefit from pretraining on in-domain data. In future work, we plan to explore the utility of decoder-only LLMs for clinical diagnostic reasoning.



\section{Limitations}

The limitation of this work was the use of ROUGE-L as the evaluation metric. Given the many acronyms and synonyms in medical writing, ROUGE-L, based on the longest common sequence, does not capture the many nuances in its score. Researchers have shown concerns for the ROUGE score and have developed metrics for summarization that are more semantically aware of the ground truth \cite{akter-etal-2022-revisiting}, but their usability is yet to be validated. 
 
Training large language models from scratch uses a considerable amount of carbon footprint. \cite{patterson2021carbon}  Fine-tuning large language models for downstream tasks is one way to reduce carbon footprint but still needs to be cost-effective. As the AI community progresses in this field, developing a cost-effective and carbon-friendly solution is needed. The NLP field is moving towards prompt-based methods with larger LLMs \cite{lester-etal-2021-power}, so the next step for this research is to experiment with soft prompting approaches to address low resource settings and leverage prompt tuning in LLMs for the problem summarization task. 

\section{Ethics Statement}


This research utilized a deidentified dataset that does not include any protected health information. This dataset operates in compliance with the PhysioNet Credential Health Data Use Agreement (v1.5.0). All experiments conducted adhered to the guidelines outlined in the PhysioNet Credentialed Health Data License Agreement. Additionally, this study has been deemed exempt from human subjects research.



\bibliography{bibliography}
\bibliographystyle{acl_natbib}

\appendix

\section{Appendix}
\label{sec:appendix}

This section adds two more tables to show the results of the other clinical task. The results are compared with previous results and can be seen in the baseline column. In many cases, such as MedNLI, AP, and EmrQA, we can see improvement in the multitask experiments.    

\begin{table*}
\begin{tabular}{l l l l l l}
\toprule
Model & Training & Summarization & SOAP & A\slash P \\
\midrule
\small \citealp{GAO2023104286} & \small Single task & \small 7.60 (5.31 - 9.89) & \small \textbf{60.12} \small (59.33 - 60.90) & \small 80.09 (79.32 - 83.23)  \\
\midrule
\small \textsc{T5 220M} & \small Single task & \small 26.35 (22.18 - 30.52) & \small \textbf{60.12} (59.33 - 60.90)* & \small 73.31 (71.34 - 77.65)*  \\
 & \small Multi-task & \small 24.84 (20.28 - 29.40) & \small 56.63 (55.83 - 57.42)& \small 43.25 (41.35 - 66.59) \\
\midrule
\small \textsc{T5 770M} & \small Single task & \small 26.90 (22.58 - 31.23) & \small 55.57 (54.78 - 56.35)* & \small 77.96 (75.38 - 81.60)* \\
 & \small Multi-task & \small 23.99 (19.86 - 28.13) & \small 51.10 (50.32 - 51.91)& \small 75.15 (71.93 - 78.19) \\
\midrule
\small \textsc{SciFive 220M} &\small Single task & \small 25.31 (21.45, 29.17) & \small 57.74 (56.95 - 58.53)* & \small 76.76 (74.81 - 80.92)* \\
 & \small Multi-task & \small 24.38 (19.99 - 28.78) & \small 54.86 (54.06 - 55.65) & \small 68.87 (65.50 - 72.12) \\
\midrule
\small \textsc{SciFive 770M} & \small Single task & \small 27.31 (23.09 - 31.53) & \small 47.65 (46.85 - 48.47)* & \small 75.11 (73.10,79.42)* \\
 & \small Multi-task & \small 25.31 (21.45 - 29.17) & \small 44.51 (43.72- 45.29)& \small 77.50 (74.45 - 80.37) \\
\midrule
\small \textsc{Clinical-T5 220M} & \small Single task & \small 25.35 (21.19 - 29.51) & \small 55.30 (54.51 - 56.11) & \small 80.44 (77.47 - 83.35) \\
 & \small Multi-task & \small 26.21 (21.92 - 30.49) & \small 52.41 (51.62 - 53.20) & \small 65.49 (62.08 - 68.76) \\
\midrule
\small \textsc{Clinical-T5 770M} & \small Single task & \small 28.28 (24.17 - 32.38) & \small 52.82 (52.03 - 53.61) & \small 78.79 (75.76 - 81.66) \\
 & \small Multi-task & \small \textbf{28.55} (24.29 - 32.80) & \small 54.00 (53.21 - 54.80) & \small \textbf{80.58} (77.57 - 83.38) \\
\hline

\end{tabular}
\caption{Finetuned T5 models on various clinical task with 95\% confidence interval calculated using the bootstrapping method. A\slash P represents assessment and plan relational labeling task. Summarization use ROUGE L, A\slash P use F1-macro and SOAP use accuracy score for the evaluation metrics. The first row in the table represents best scores reported in the DR.BENCH paper and * in the other rows represent scores for the respective task in DR.BENCH paper \cite{GAO2023104286}}
\label{tab:submission3}
\end{table*}

\begin{table*}
\begin{tabular}{l l l l l l}

\toprule
Model & Training & EmrQA & MedNLI & MedQA \\
\midrule
\small \citealp{GAO2023104286} & \small Single task & \small 39.20 (34.63 - 43.78) & \small 84.88 (82.98 - 86.64) & \small 24.59 (22.31 - 27.02)  \\
\midrule
\small \textsc{T5 220M} & \small Single task & \small 33.40 (29.27 - 37.61)* & \small 79.75 (78.62 - 82.70)* & \small 22.55 (20.01 - 25.69)* \\
 & \small Multi-task & \small 38.48 (37.24 - 39.79) & \small 72.57 (70.18 - 74.82) & \small 21.75 (19.48 - 24.12) \\
\midrule
\small \textsc{T5 770M} & \small Single task & \small 38.05 (33.56 - 42.58)* & \small 84.04 (82.14 - 85.86)* & \small 20.97 (18.77 - 23.25)* \\
 & \small Multi-task & \small 41.42 (40.16, 42.72) & \small 83.19 (81.22, 85.09) & \small 23.25 (20.97, 25.61) \\
\midrule
\small \textsc{SciFive 220M} &\small Single task & \small 37.28 (32.84 - 42.11)* & \small 82.84 (80.87 - 84.74)* & \small 22.78 (20.50 - 25.14)* \\
 &\small Multi-task & \small 40.08 (38.82 - 41.39) & \small 78.83 (76.72 - 80.94) & \small 21.52 (19.32 - 23.80) \\
\midrule
\small \textsc{SciFive 770M} & \small Single task & \small 41.21 (39.93 - 42.49) & \small 83.89 (82.00 - 85.79) & \small 23.09 (20.82 - 25.37) \\
 & \small Multitask & \small 41.26 (39.98 - 42.56) & \small 84.35 (82.49 - 86.22) & \small 23.72 (21.37 - 26.08) \\
\midrule
\small \textsc{Clinical-T5 220M} & \small Single task & \small 41.35 (40.07 - 42.65) & \small 84.32 (82.42 - 86.15) & \small 21.92 (19.64 - 24.19) \\
 & \small Multi-task & \small 40.30 (39.02 - 41.62)& \small 71.23 (68.92 - 73.56)& \small 22.46 (20.19 - 24.74) \\
\midrule
\small \textsc{Clinical-T5 770M} & \small Single task & \small \textbf{42.69} (41.39 - 43.95)& \small 85.86 (85.02 - 88.47)& \small 24.27 (21.92 - 26.63) \\
 & \small Multi-task & \small 42.61 (41.34 - 43.92)& \small \textbf{86.14} (84.32 - 87.90)& \small \textbf{25.84} (23.41 - 28.28) \\
\bottomrule
\end{tabular}
\caption{Finetuned T5 models on various clinical task with 95\% confidence interval calculate using the bootstrapping method. All the evaluation metrics here are the accuracy score. The first row in the table represents best scores reported in the DR.BENCH paper and * in the other rows represent scores for the respective task in DR.BENCH paper \cite{GAO2023104286}}
\label{tab:submission4}

\end{table*}

\begin{table*}
\centering
\begin{tabular}{p{5.7cm} p{2.1cm} p{2.1cm} p{2.1cm} p{2.1cm}}

\toprule
\small \textbf{Input} & \small \textbf{Ground Truth} \newline
\textbf{Diagnoses}\slash \textbf{Problems} & \small \textbf{T5 770M} \newline
\textbf{Single task} & \small \textbf{Clinical-T5} \newline
\textbf{770M} \newline
\textbf{Single task} & \small \textbf{Clinical-T5} \newline
\textbf{770M} \newline
\textbf{Multi-task}  \\
\midrule
\small SUMMARIZE:  \textit{\textbf{<ASSESSMENT>}} 48 y/o M \newline 
with HIV 47M s/p elective spinal surgery \newline 
(anterior and posterior LIFs),
intubated - - - -  \textit{\textbf{<SUBJECTIVE>}} Agitated, diaphoretic, \newline
altered, hypertensive and tachy this AM - - - - \textit{\textbf{<OBJECTIVE>}} Last dose of Antibiotics: \newline
Infusions:
Other ICU medications:
Heparin \newline
Sodium (Prophylaxis) - - - - 
& 
\small
EtOH \newline 
withdrawal Spinal surgery 
& 
\small
Altered MS \newline
s p elective \newline 
spinal surgery 
& 
\small
DTs EtOH w d \newline 
pain h o chronic pain 
& 
\small
Altered mental \newline
status \newline 
Hypertension Tachycardia Acute renal \newline
failure s p spinal \newline
surgery \\

\midrule
\small SUMMARIZE:  \textit{\textbf{<ASSESSMENT>}} SEPSIS \newline
WITHOUT ORGAN DYSFUNCTION Ms. \newline
[**Known lastname 10381**] is a 76F with \newline
multiple medical problems, who is - - - - - - - \newline
\textit{\textbf{<SUBJECTIVE>}} FEVER - 101.7 F - [**2129-9-3**] 12:33 PM
-received boluses overnight \newline
for low SBP
- - - -   \textit{\textbf{<OBJECTIVE>}} Last dose of 
Antibiotics: Cefipime - [**2129-9-3**] 04:05 PM
Metronidazole - [**2129-9-4**] 04:00 AM - - - -
& 
\small
Sepsis Patient \newline
has re developed \newline
fevers on 9 2 on \newline
a regimen of \newline
vancomycin ceftriaxone Possible sources \newline
include 1 Intra \newline 
abdominal source 
& 
\small
Sepsis Thrombocytopenia
& 
\small
Sepsis Likely \newline 
source is \newline 
clostridium \newline
difficile colitis \newline 
Acute renal \newline
failure 
& 
\small
Hypotension Likely \newline
secondary to \newline 
sepsis though \newline 
source unclear at this time Acute \newline 
renal failure \\
\bottomrule
\end{tabular}
\vspace{-.1in}
\caption{The table represents a snippet of the input and output sections of problem summarization. The input data contains an added prefix that denotes the task for T5, "SUMMARIZE" in this case, and <prefix> that defines the note section. Finally, "- - - -" is the continuation of the section, which was excluded here due to the space constraint.}
\label{tab:erroranalysissnippit}
\end{table*}

\end{document}